\documentclass[runningheads]{llncs}
\usepackage{graphicx}

\usepackage{tikz}
\usepackage{comment}
\usepackage{amssymb} 
\usepackage{amsmath}
\usepackage{bm}
\usepackage{color}
\usepackage{enumitem}
\usepackage{algorithm}
\usepackage{algorithmic}
\usepackage{epsfig}
\usepackage{graphicx}
\DeclareMathOperator*{\argmin}{argmin}
\DeclareMathOperator*{\argmax}{argmax}

\usepackage{lipsum}
\newcommand\blfootnote[1]{%
  \begingroup
  \renewcommand\thefootnote{}\footnote{#1}%
  \addtocounter{footnote}{-1}%
  \endgroup
}

\begin{document}
\pagestyle{headings}
\mainmatter

\title{Data-Free Neural Architecture Search via Recursive Label Calibration}

\titlerunning{Data-free NAS}

\author{Zechun Liu\inst{1,2} \and
Zhiqiang Shen\inst{2,3}\thanks{Corresponding author.} \and
Yun Long\inst{4} \and
Eric Xing\inst{2,3} \and
Kwang-Ting Cheng\inst{1}\and
Chas Leichner\inst{4}}
\authorrunning{Zechun Liu et al.}
\institute{Hong Kong University of Science and Technology \and
Carnegie Mellon University \and
Mohamed bin Zayed University of Artificial Intelligence \and
Google Research\\}
\maketitle

\begin{abstract}
This paper aims to explore the feasibility of neural architecture search (NAS) given only a pre-trained model without using any original training data. This is an important circumstance for privacy protection, bias avoidance, etc., in real-world scenarios. To achieve this, we start by synthesizing usable data through recovering the knowledge from a pre-trained deep neural network. Then we use the synthesized data and their predicted soft-labels to guide neural architecture search. We identify that the NAS task requires the synthesized data (we target at image domain here) with enough semantics, diversity, and a minimal domain gap from the natural images. For semantics, we propose recursive label calibration to produce more informative outputs. For diversity, we propose a regional update strategy to generate more diverse and semantically-enriched synthetic data. For minimal domain gap, we use input and feature-level regularization to mimic the original data distribution in latent space. We instantiate our proposed framework with three popular NAS algorithms: DARTS~\cite{liu2018darts}, ProxylessNAS~\cite{cai2018proxylessnas} and SPOS~\cite{guo2019single}. Surprisingly, our results demonstrate that the architectures discovered by searching with our synthetic data achieve accuracy that is comparable to, or even higher than, architectures discovered by searching from the original ones, for the first time, deriving the conclusion that NAS can be done effectively with no need of access to the original or called natural data if the synthesis method is well designed. \blfootnote{This work is done when Zechun Liu is an intern at Google Research.}
\end{abstract}

\section{Introduction}

Neural architecture search (NAS) has demonstrated substantial success in automating the design of neural networks~\cite{zoph2018learning,baker2016designing,real2017large,real2018regularized,liu2018progressive,tan2018mnasnet,jin2018efficient}.
A typical NAS algorithm usually involves three core components: a search space, a search algorithm, and a set of training data. The majority of researches in NAS focuses on search space design~\cite{liu2018darts,ying2019bench} or exploring superior search algorithms~\cite{cai2018proxylessnas,wu2018fbnet}. While, in this study, we investigate the feasibility of performing neural architecture search without accessing the original training data.
We assume that we only have a pre-trained model, and the original dataset is not accessible during the neural architecture search process. This is a common and useful circumstance that is needed to be solved urgently for privacy protection, bias avoidance, etc. We call this \textit{data-free NAS}, a practical task for application scenarios in which privacy or logistical concerns restrict sharing of the original training data but permit sharing of a model trained by such data for NAS. Also, models are usually smaller in size than large-scale datasets, which makes them easier to exchange and store.

Conducting NAS without data is challenging. Traditional NAS relies on the input images to train and rank different architectures. A natural image dataset contains semantic patterns and inter-class relationships, which are helpful in guiding architecture search. For scenarios where the original data is not accessible, we first need to synthesize an image dataset from the model pre-trained on the original data, and use such a synthesized dataset for conducting architecture search. This, however, raises a crucial question: how do we synthesize an image dataset that has the important search-relevant attributes or properties of the original data for effective NAS without using any of the original data?

Currently, data-free NAS is an under-explored task that requires unique understanding of the particular synthesized data. In this work, we empirically identify three attributes that NAS in data-free scenario requires the synthesized images to have: (i) rich semantic information, (ii) sufficient image diversity, and (iii) a minimized domain gap with the original data. Rich semantic information ensures the classification task on the synthesized images is as complex as that of the original training images. High image diversity prevents the NAS algorithms from overfitting to the synthesized data and producing trivial solutions. A minimized domain gap makes certain that architectures found by searching over the synthesized data also can perform well on the original data.

\begin{figure*}[t]
    \centering
    \includegraphics[width=0.95\linewidth]{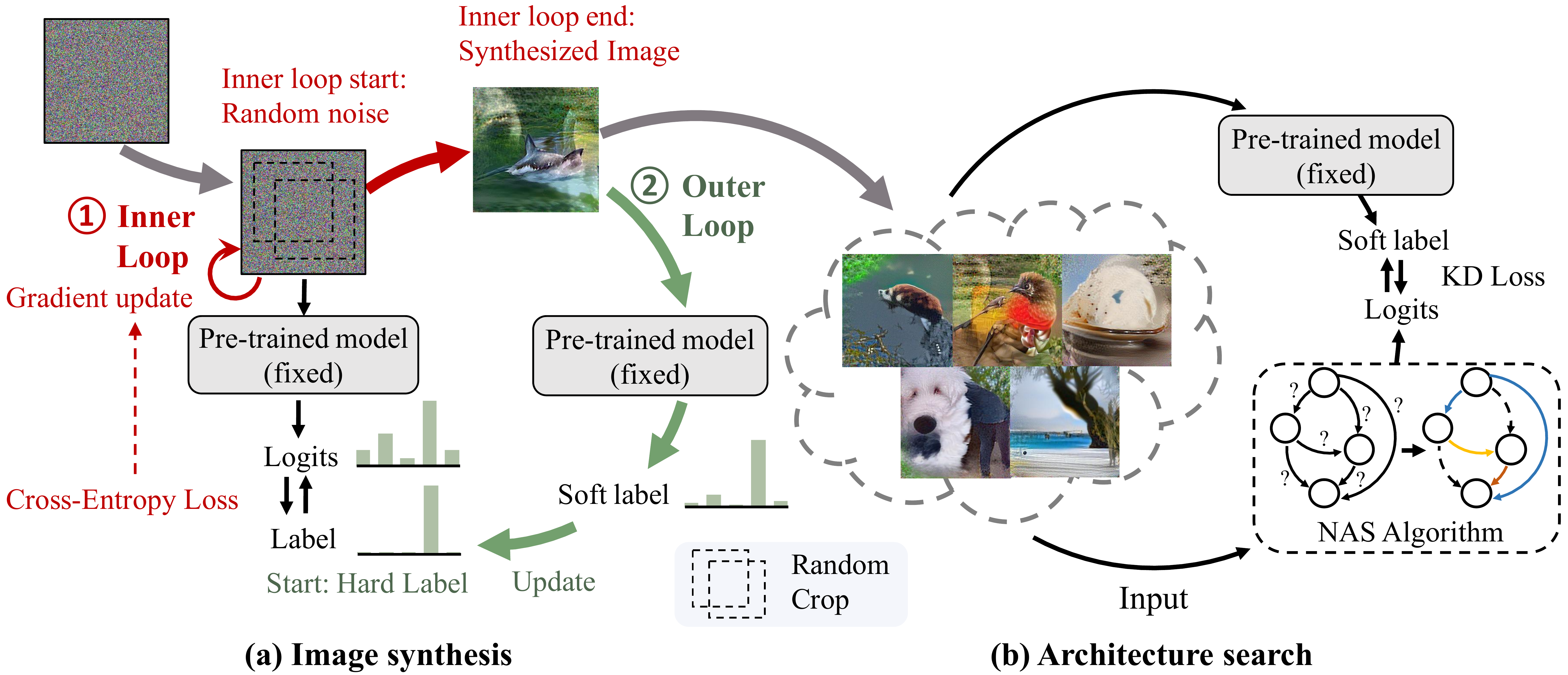}
    \vspace{-0.12in}
    \caption{The proposed data-free NAS framework consists of two stages: (a) image synthesis using recursive label calibration, and regional update; (b) running an architecture search using the synthesized images. Specifically, the image synthesis stage consists of an \textcolor[RGB]{176,35,24}{\textbf{Inner Loop}} and an \textcolor[RGB]{36,82,18}{\textbf{Outer Loop}}. The \textcolor[RGB]{176,35,24}{\textbf{Inner Loop}} starts with a random noise input, which is then updated with the gradient computed from the cross-entropy loss (CE-loss). The inner loop ends when random noise converges to a batch of synthesized images. Then in the \textcolor[RGB]{36,82,18}{\textbf{Outer Loop}}, we infer the predictions of the synthesized images using the pre-trained model and use these predictions as the soft labels for synthesizing the next batch of images. See section~\ref{sec:method} for details.}
    \vspace{-0.15in}
    \label{fig:overall}
\end{figure*}

To fulfill these requirements, we propose a novel image synthesis method for promoting the effectiveness of architecture search procedure. As shown in Fig.~\ref{fig:overall}, we synthesize the images using gradient descent with respect to class labels. We observe that conducting NAS on data synthesized from one-hot labels leads to overfitting and the models searched by the NAS algorithms fail to generalize to the original training data. This issue is caused by one-hot labels' limitation of being unable to capture the full set of semantic relationships between classes in the original training data. For instance, the synthesized images will be confidently classified as a single class (\textit{e.g.}, ``coffee mug'') but with no trace of similarity to other relevant classes (\textit{e.g.}, ``cup''). To avoid this, we propose \textit{recursive label calibration} for finding semantically-significant soft labels.

As shown in Fig.~\ref{fig:overall} (a), we use logits derived from the pre-trained model as the class labels during the image synthesis process. Starting from a hard label, i.e., one-hot distribution, we recursively apply this labeling process to successive batches of images to amplify the representation of classes which are semantically related to the original target class while only weakly presented in the initial image synthesis. We show that images synthesized using recursive label calibration are more diverse and can better capture semantic relationships that exist in the original training data. In order to further improve the diversity, we propose a novel regional gradient updating scheme on synthesized images to match the random crop augmentation in training NAS.

We show that, the architecture accuracy rankings on the data synthesized with the proposed method produce more consistent correlation with the rankings on the original data. In turn, NAS on our synthesized data discovers the searched architecture with much higher accuracy. We demonstrate the feasibility of our data-free NAS conversion on three prevalent NAS algorithms: a reinforcement-learning-based algorithm (ProxylessNAS~\cite{cai2018proxylessnas}), an evolution-based algorithm (Single Path One-Shot~\cite{guo2019single}) and a gradient-based algorithm (DARTS~\cite{liu2018darts}).

\vspace{0.2em}
\noindent We make four major contributions in this paper:
\vspace{-0.6em}
\begin{itemize}[leftmargin=*]\setlength{\itemsep}{-2pt}
\item We reveal, for the first time, that it is feasible to search architectures without relying on the original training data and propose a framework for data-free NAS.
\item We identify properties of synthesized images that NAS requires, and propose a data synthesis method that uses recursive label calibration and regional update to generate images with sufficiently high diversity from the information stored in a pre-trained model for effective NAS.
\item We validate the generalization ability of data-free NAS by integrating it with three different NAS algorithms and demonstrating competitive results.
\item We further extend the scope of our data synthesis method and demonstrate that it also outperforms prior approaches for the data-free pruning and knowledge transfer tasks.
\end{itemize}

\section{Related Work}

{\bf Generative adversarial networks (GAN)} \cite{brock2018large,zhu2017unpaired} can generate images with high fidelity, but still need real images as a reference while training the generator, which is not fully data-free. Recently, Chen et al.~\cite{chen2019data} and Xu et al.~\cite{xu2020generative} proposed to use a generator to synthesize images from a pre-trained model and simultaneously train the student network. Further, Yin et al.~\cite{yin2020dreaming} proposed to synthesize images from the pre-trained teacher network using regularization terms and Jensen-Shannon divergence loss. However, such method is a general design without explicitly considering the requirements for architecture search and also ignoring the crucial factors like the semantic diversity in the latent feature and image spaces. In this work, we propose recursive label calibration and regional update to generate more diverse and semantically meaningful images to fulfill the demand of neural architecture search.
{\bf NAS} is a tool for automatic discovery of optimized neural network architectures under various practical constraints. To conduct NAS, it normally requires a search space, a search algorithm and a set of training data. Current NAS research mainly focuses on improving the search algorithms~\cite{real2018regularized,zoph2016neural,baker2016designing,zoph2018learning}, designing the search space~\cite{real2017large,dai2018chamnet,ying2019bench}, reducing the search cost~\cite{brock2017smash,jin2018efficient,pham2018efficient,cai2018proxylessnas,liu2018darts} and integrating direct metrics with the search process~\cite{wu2018fbnet,guo2019single}.
To our best knowledge, there is no previous literature directly studying the problem of data-free NAS. Thus, our work has many practical and useful guidelines on this problem for future research. One prior work~\cite{liu2020labels} may be the closest one to ours, it proposed to conduct NAS without human-annotated labels. In our study, we take one step further and attempt to answer the question of whether NAS can be conducted even without the original data at all. This work focuses on investigating the properties of the original data that NAS algorithm depends on.

\section{Our Method}
\label{sec:method}

In data-free neural architecture search, as we have no access to the original data $\mathbf{X}\in \mathbb{R}^{w \times h}$ , we aim to use synthesized data $\mathbf{\hat{X}}\in \mathbb{R}^{w \times h}$ to search for the high-performing architecture $\mathit{\mathbf{A}}^*$ in the search space of $\mathcal{S}$, with the NAS algorithm minimizing the loss:
\begin{align}
\mathit{\mathbf{A}}^* = \argmin_{\mathbf{A} \in \mathcal{S}} \  \mathcal{L}(\mathbf{A}|\mathbf{\hat{X}}).
\label{eq:problem_setup}
\end{align}
In this way, we target at finding an architecture that achieves high performance when evaluated on the target data $\mathcal{X}$:
\begin{align}
\mathbf{A}^* \approx \argmax_{\mathbf{A} \in \mathcal{S}}  \  \rm{\mathbf{Acc}}_{eval}(\mathbf{A}|\mathbf{X}).
\label{eq:problem_setup2}
\end{align}
This requires the synthesized images to be semantically meaningful, diverse, and have a minimum domain gap with the target data (\textit{i.e.}, $\mathbf{\hat{X}} \sim \mathbf{X}$) to ensure that the architecture rankings on the synthesized images align well with the rankings on original data. Such that we can conduct data-free NAS by integrating the synthesized data and their corresponding soft labels with existing NAS algorithms.

\vspace{-0.1in}
\subsection{Data Synthesis}
\label{sec:data_synthesis}
We optimize the classification (cross-entropy) loss between the pre-trained model's prediction $\mathbf{M}_{\rm{pretrained}}(\hat{\mathbf{X}})$ on the synthesized data and the target label $y$:
\begin{align}
\min_{\hat{\mathbf{X}}} \mathcal{L}_{CE}(\mathbf{M}_{\mathbf{pretrained}}(\hat{\mathbf{X}}),\mathbf{y}),
\label{eq:loss_CE}
\end{align}
where $\mathcal{L}_{CE}$ denotes the classification loss (\textit{i.e.}, cross-entropy loss) and $\hat{\mathbf{X}}$ denotes input ``image''.

Specifically, since one-hot label $\mathbf{y}$ is unable to capture the underlying class relationship, we propose recursive label calibration to automatically learn the semantically-related soft labels $\hat{\mathbf{y}}$ and use $\hat{\mathbf{y}}$ for image synthesis. Furthermore, we propose a novel regional update scheme to improve the diversity of generated images.

\noindent{\textbf{Regularization}}

\noindent (i) \textit{Input-level Regularization:}

As a natural image taken from a real scene is unlikely to sharply vary in value between adjacent pixels, following~\cite{mordvintsev2015inceptionism}, we impose a regularization term $\mathcal{L}_{\bm {input}}$ to penalize the overall variance in the synthesized data:
\begin{equation}
\small{
\mathcal{L}_{\bm{input}}=\sum_{\bm{i,j}}((\hat{\mathbf{X}}_{\bm{(i+1,j)}}- \hat{\mathbf{X}}_{\bm{(i,j)}})^2+(\hat{\mathbf{X}}_{\bm{(i,j+1)}}-\hat{\mathbf{X}}_{\bm{(i,j)}})^2)
}
\label{eq:loss_input}
\end{equation}
\noindent where \textit{i, j} denotes the index of each pixel in the image.

\noindent (ii) \textit{Feature-level Regularization:}

Further, we impose a regularization term to align the high-level feature map statistics with the target images by enforcing the synthesized images to produce similar statistics in the feature maps as statistics calculated and stored in each BN layer of the pre-trained model, as~\cite{yin2020dreaming}.
\begin{equation}
\small{
\mathcal{L}_{\bm{feat}}=\sum_{\bm{l}}(||\mathbf{E}_{\bm{l}}(\hspace{-0.08em}\hat{\mathbf{X}}\hspace{-0.08em})-\mathbf{E}_{\bm{l}}(\hspace{-0.08em}\mathbf{X}\hspace{-0.08em})||_2+|| \mathbf{Var}_{\bm{l}}(\hspace{-0.08em}\hat{\mathbf{X}}\hspace{-0.08em})\hspace{-0.08em}-\hspace{-0.08em}\mathbf{Var}_{\bm{l}}(\hspace{-0.08em}\mathbf{X}\hspace{-0.08em})||_2)}
\label{eq:loss_feature}
\end{equation}
Here,  ${\rm E}_l(\hat{\mathbf{X}})$ and ${\rm Var}_l(\hat{\mathbf{X}})$ denote the mean and variance of the feature map of synthesized images in the $l^{th}$ convolutional layer, ${\rm E}_l(\mathbf{X})$ and ${\rm Var}_l(\mathbf{X})$ denote the historical mean and variance of target images, which are stored in BatchNorm~\cite{ioffe2015batch} of the pre-trained model.

\vspace{-0.06in}
\subsubsection{Recursive Label Calibration}
\label{sec:recursive_label_calibration}
\begin{figure}[t]
    \centering
    \includegraphics[width=0.75\linewidth]{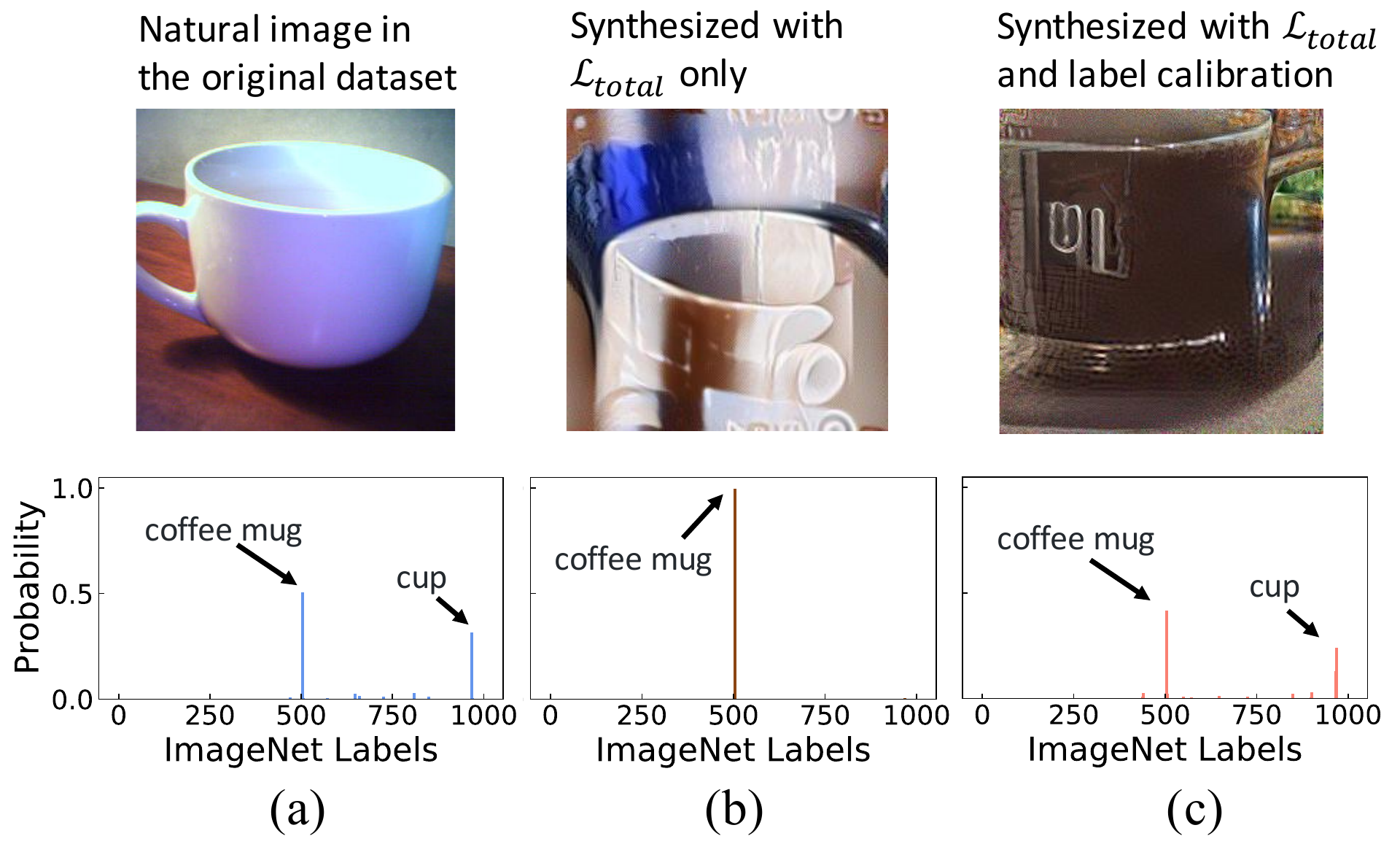}
    \vspace{-0.1in}
    \caption{The pre-trained model's prediction on the input image.
    (a) Natural image in the category \textit{``coffee mug''}. Its prediction contains a peak at \textit{``coffee mug''} and another peak value in the related class \textit{``cup''}.
    (b) The image synthesized \textit{w.r.t.} the \textit{``coffee mug''} class as a one-hot hard label. The pre-trained model only strongly predicts this image as \textit{``coffee mug''}. Meanwhile, there is a pattern on the surface of the mug that resembles coffee.
    (c) The image synthesized using recursive label calibration. It automatically identifies \textit{``cup''} as a related class for the image synthesized \textit{w.r.t.} the \textit{``coffee mug''} category. The image looks more natural as well.}
    \label{fig:label_calibration}
\end{figure}

Importantly, we observe that the pre-trained model's prediction on a natural image will spread as a distribution of logits, with the maximum value being the target class and other several peaks landing at similar classes, shown in Fig.~\ref{fig:label_calibration} (a).

\begin{figure}[t]
    \centering
    \includegraphics[width=0.8\linewidth]{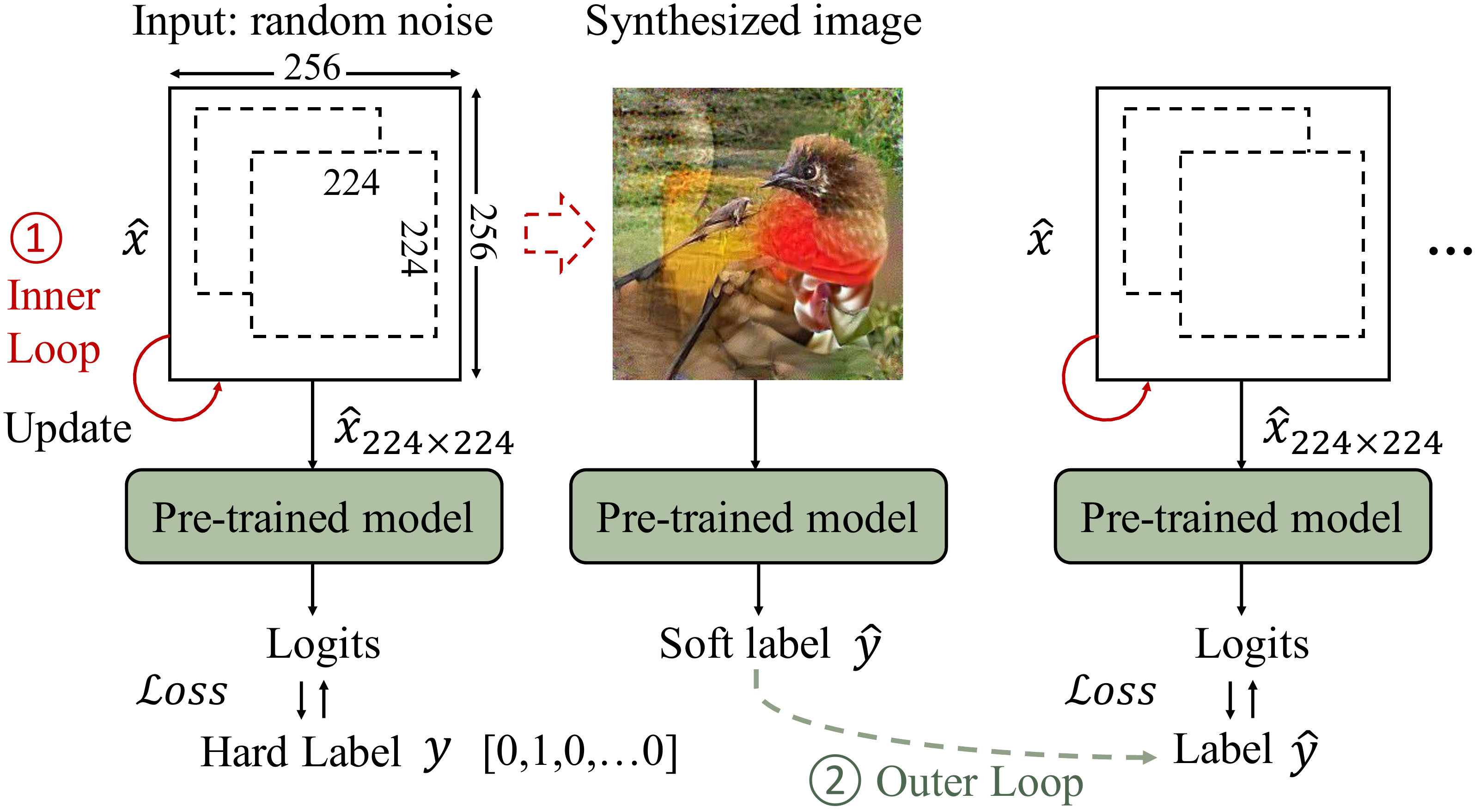}
    \vspace{-0.5em}
    \caption{An illustration of the proposed regional update scheme and recursive label calibration for image synthesis. In every iteration of the generation process, a 224$\times$224 region is randomly cropped from the 256$\times$256 input and we update the selected region using gradients calculated \textit{w.r.t.} Eq.~\ref{eq:loss_CE}. After an image is synthesized, we use the pre-trained model's prediction on the synthesized image as the soft label for computing the cross-entropy loss with the output logits of the next image batch.}
    \label{fig:detail}
    \vspace{-0.1in}
\end{figure}

Thus, for encouraging synthesized images to model higher-level class correlation and to capture more subtle semantic information, we want the targets of synthesized images to be a distribution of semantically-related classes. Considering no original images are available, we cannot obtain the relationship between classes directly. However, we make an important observation that the pre-trained model's prediction on the image synthesized with respect to the hard label also has fractional logits spread between semantically similar classes. But these logits are too weak to be reflected in the synthetic images. Thus, we propose recursive label calibration to amplify this distribution, which utilizes the soft prediction of the previously synthesized image as the new targets,
\begin{align}
\hat{\mathbf{y}}_{\mathbf{t}\mathbf{-1}} = \mathbf{M}_{\rm{\mathbf{pretrained}}}(\hat{\mathbf{X}}_{\mathbf{t}-1}),
\label{eq:soft_target}
\end{align}
replacing the hard label $y$ in Eq.~\ref{eq:loss_CE} to guide the image synthesizing process:
\begin{align}
\hat{\mathbf{X}}_t^* = \min_{\hat{\mathbf{X}}} \mathcal{L}_{CE}(\mathbf{M}_{\mathbf{pretrained}}(\hat{\textbf{X}}_t),\hat{\mathbf{y}}_{t-1})
\label{eq:soft_image}
\end{align}
as also illustrated in Fig.~\ref{fig:detail}.

After visualizing the images synthesized with recursive label calibration, we confirm that these images do learn the semantic relationships among classes. As shown in Fig.~\ref{fig:label_calibration} (b), the image synthesized without label calibration is overconfident at the prediction of the ``coffee mug'' class. While in Fig.~\ref{fig:label_calibration} (c), the prediction of the image synthesized with recursive label calibration automatically learns that ``coffee mug'' and ``cup'' are two related classes. Recursive label calibration encourages the synthesized images to learn class relationships as natural images instead of over-fitting to a single class, which in turn allows the synthesized images to encode more semantic information.

In addition, the prediction of the most related classes differs between synthetic image batches due to stochasticity in the synthesis process. Since each synthesis target is now a weighted combination of related classes rather than a single class, the number of distinct targets is greatly increased. More targets lead to more diversity in the images synthesized from these targets and produce a synthesized dataset that better resembles the natural training set. We show that higher similarity between synthetic and natural dataset leads to consistency in evaluation accuracy between architectures trained on original data and the synthetic data.

\subsubsection{Regional Update Scheme}

To further enhance synthesized image diversity, we propose a regional update scheme for synthesis process. The regional update is formulated as:
\begin{align}
\mathbf{G}_{\hat{\mathbf{X}}}\!=\!\nabla_{\hat{\mathbf{X}}}\mathcal{L}_{CE}(\mathbf{M}_{\mathbf{pretrained}}(\hat{\mathbf{X}}_{w, h \in \textbf{R}_{\mathbf{selected}}}),\mathbf{y}),
\label{eq:loss_regional}
\end{align}
Instead of generating the images with the size required for forward computation in the neural network, \textit{e.g.}, 224$\times$ 224 for ImageNet dataset, we enlarge the size of the input tensor $\hat{\mathbf{X}}$, \textit{e.g.}, 256$\times$256 for ImageNet dataset. In every iteration, we randomly select a sub-region $\textbf{R}_{\mathbf{selected}}$, \textit{e.g.}, 224$\times$224 as the input to the pre-trained model and calculate the gradients $\mathbf{G}_{\hat{\mathbf{X}}}$. We only update the selected region, leaving the pixels outside of that region unchanged, as also shown in Fig.~\ref{fig:detail}.
This proposed regional update during image synthesis reflects the random crop data augmentation during training, which greatly increases the amount of trainable data and enhances the image diversity. Additionally, it can encourage the translation-invariance in the synthesized images.

\begin{figure}[t]
    \centering
    \includegraphics[width=0.72\linewidth]{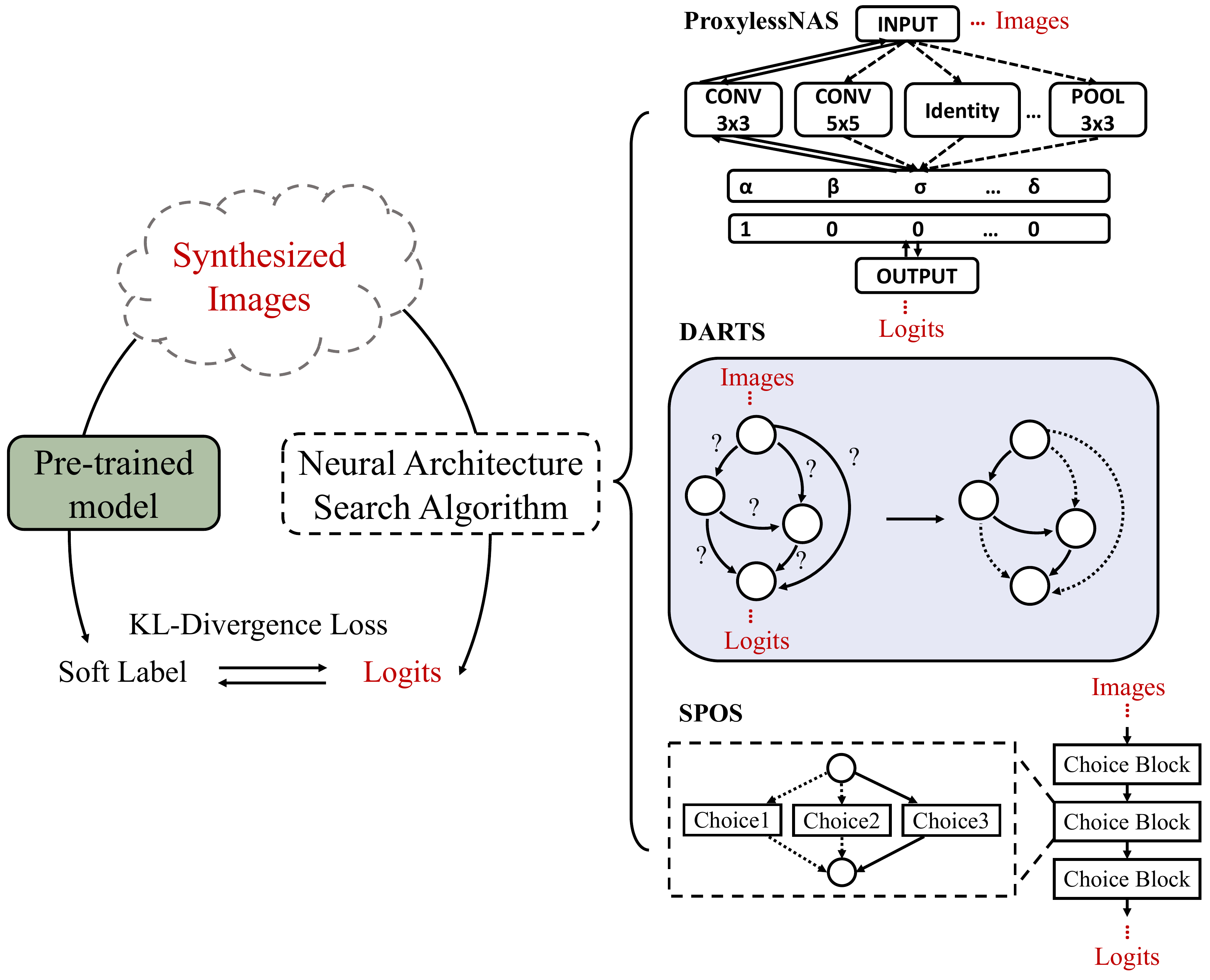}
    \vspace{-1em}
    \caption{The proposed data-free neural architecture search framework using the synthesized images. We replace the original ones with the synthesized images and use the soft labels to compute the KL-divergence loss with the output logits of NAS.}
    \label{fig:integrate}
\end{figure}

\vspace{-0.06in}
\subsection{Architecture Search}
After extracting the classification-relevant information from the pre-trained model by constructing a synthetic dataset, we use that dataset to conduct neural architecture search. Fig.~\ref{fig:integrate} illustrates the proposed data-free neural architecture search framework: data-free NAS can be applied to an existing NAS algorithm by replacing the natural training data with synthetic data, and the classification labels with soft-labels from pre-trained model. We instantiate data-free NAS with three prevalent NAS algorithms: ProxylessNAS~\cite{cai2018proxylessnas}, DARTS~\cite{liu2018darts} and SPOS~\cite{guo2019single}, which are reinforcement-learning-based, gradient-based and evolution-based respectively:

The gradient-based neural architecture search jointly learn the architecture parameter $\bm{\alpha}$ and the SuperNet weights $\mathbf{W}$, aiming to find $\bm{\alpha}^*$
that minimizes the validation loss $\mathcal{L}_{val}$, with weights $\mathbf{W}^*$ in the SuperNet obtained by minimizing the training loss $\mathcal{L}_{train}$~\cite{liu2018darts}.
\begin{align}
    \min_{\bm{\alpha}} \ \ \ & \mathcal{L}_{val}(\mathbf{W}^*(\bm{\alpha}), \bm{\alpha}) \\
    \mathbf{s.t.} \ \ \ & \mathbf{W}^*(\bm{\alpha}) = \argmin_{\mathbf{W}}\mathcal{L}_{train}(\mathbf{W}, \bm{\alpha})
\end{align}
The reinforcement-learning-based method use the policy gradient to update the architecture parameters $\mathbf{g}$ to maximize the reward $R$~\cite{cai2018proxylessnas}.
\begin{align}
\mathcal{L}(\bm{\alpha}) & = \mathbb{E}_{\mathbf{g}\sim\bm{\alpha}}[R(\mathbf{A}_{\mathbf{g}})], \\
\nabla_{\bm{\alpha}} \mathcal{L}(\bm{\alpha}) & = \mathbb{E}_{\mathbf{g}\sim\bm{\alpha}}[R(\mathbf{A}_{\mathbf{g}}) \nabla_{\bm{\alpha}} \log(\mathbf{p}(\mathbf{g}))],
\end{align}
where $\mathbf{g}$ denotes the binary gates of choosing certain masks with the probability $\mathbf{p}$.

The evolution-based search in SPOS disentangled SuperNet training and architectural parameter optimization. Specifically for SuperNet training:
\begin{align}
    \mathbf{W}^*(\mathbf{\bm{\alpha}}) = \argmin_{\mathbf{W}}, \mathbb{E}_{\bm{\alpha}\sim\Gamma(\mathbf{A})} \mathcal{L}_{train}(\mathbf{W}, \bm{\alpha}),
\end{align}
where $\mathbf{W}$ denotes the SuperNet weights, $\bm{\alpha}$ denotes the architecture parameters and $\Gamma(\mathbf{A})$ is the probability distribution of architecture sampling.
After the supernet is trained, the weights can be used for evolutionary search in the second separate step.

In the search phase, data-free NAS explores the pre-defined search space using the same search algorithm as the NAS algorithm it integrates. Instead of using the original data for ranking different architectures, we show that data-free NAS can reliably estimate the ranking with synthetic data generated with the proposed method. Further, we show that these architectural rankings are consistent with rankings on the original data, meaning that data-free NAS algorithms can discover architectures that achieve high accuracy when evaluated on the target data.

\section{Experiments}
\label{sec:experiments}
To verify the effectiveness of our proposed data synthesis method for NAS, we first conduct consistency experiments, showing that with the proposed data synthesis method, the synthesized dataset possesses high correlation with the original data. Then, with these synthetic data, we demonstrate the feasibility of data-free NAS in discovering the architectures that perform competitively when evaluated on the target data. Further, we show that our synthesized data is also helpful in enhancing the accuracy of other tasks like data-free pruning and knowledge transfer and outperforms previous works. Lastly, we visualize the synthetic data in Sec. {\em Visualization}, and find that the proposed recursive label calibration and regional update scheme can largely improve the semantic diversity of the synthetic data and helps capture the underlying class relationships.

\begin{figure*}[t]
     \centering
     \includegraphics[width=0.9\linewidth]{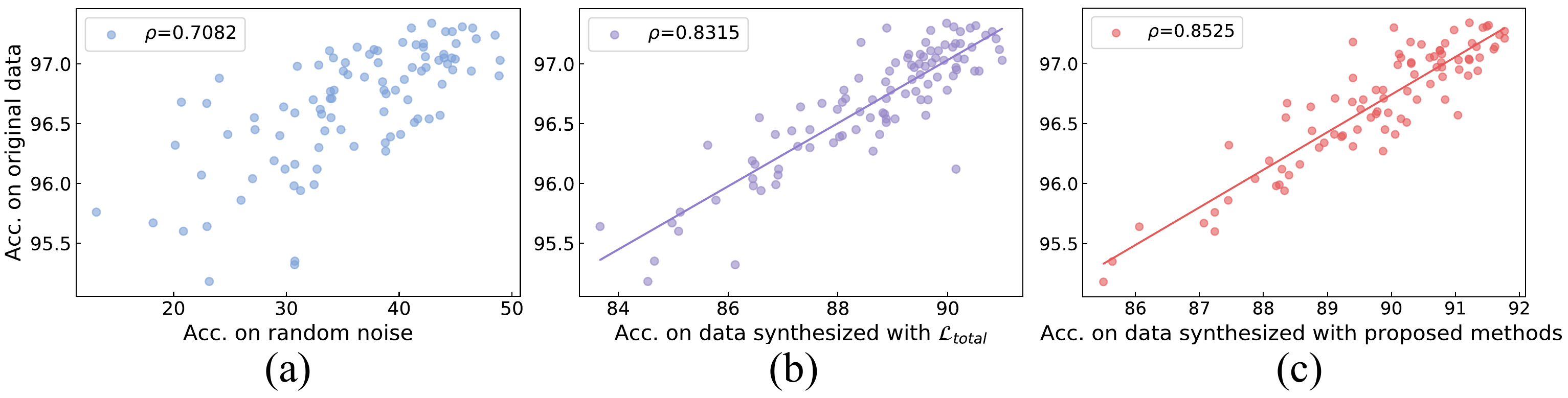}
     \vspace{-1em}
     \caption{Correlation between the accuracy of the architectures sampled from the DARTS~\cite{liu2018darts} search space on the original CIFAR-10 data and different sources of synthetic data (including random noise). Here, $\rho$ denotes Spearman’s rank correlation.}
     \label{fig:consistency_cifar10}
\end{figure*}

\begin{figure*}[t]
     \centering
     \includegraphics[width=\linewidth]{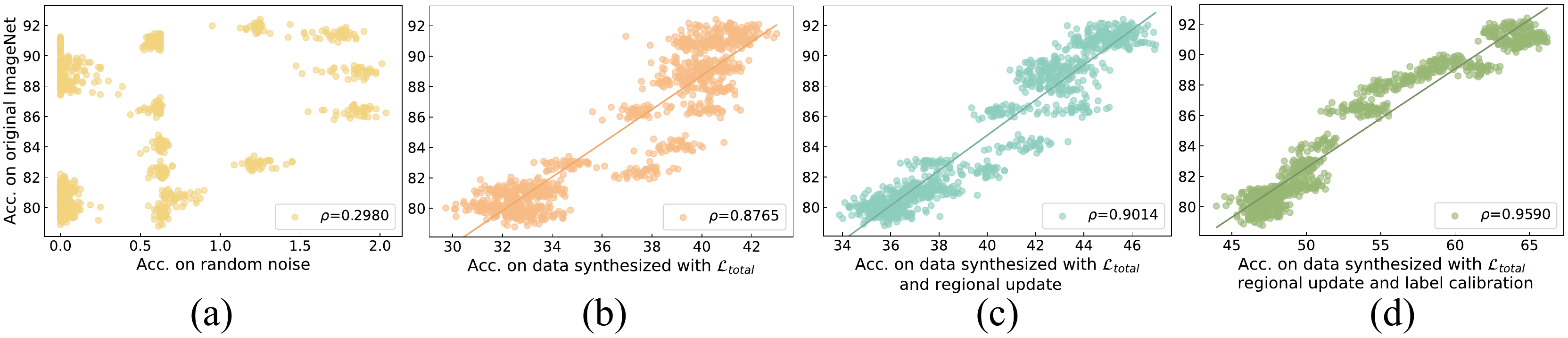}
     \vspace{-2.0em}
     \caption{Accuracy correlation on the SPOS~\cite{guo2019single} search space between original ImageNet \textit{vs.} random noise as well as data synthesized with different methods. $\rho$ denotes Spearman’s rank correlation.}
     \label{fig:consistency_imagenet}
\end{figure*}

\subsection{Consistency Exploration}
\label{sec:consistency_exp}
We conduct the consistency experiments on two search spaces: the DARTS search space~\cite{liu2018darts}, and the Single Path One-Shot (SPOS) search space~\cite{guo2019single}. The term \textit{consistency} refers to the accuracy correlation between the architectures trained on the synthetic data and the original data.

\subsubsection{DARTS Search Space on CIFAR-10 Dataset}

Fig.~\ref{fig:consistency_cifar10} shows the correlation between the accuracy of architectures trained on synthetic data and the original data. Spearman’s Rank Correlation~\cite{spearman1961proof}, denoted as ``$\rho$'', is calculated in order to quantify the correlations between accuracy on synthesized and original data. From Fig.~\ref{fig:consistency_cifar10} (a) we can see that the overall accuracy of the network trained on the random noise is lower than 50\% and the ranking is highly noisy, the random noise can hardly guide NAS for finding a high-quality architecture. In contrast, synthesized data improves the correlation compared to random noise and recursive label calibration further enhances the correlation. This correlation supports our belief that a NAS algorithm applied on the good synthetic data is likely to discover architectures that perform well on the original data.

\vspace{-0.06in}
\subsubsection{SPOS Search Space on ImageNet dataset}

As shown in Fig.~\ref{fig:consistency_imagenet} (a), for the large-scale ImageNet, simply using the random noise results in a totally uncorrelated accuracy between original and the ``noise data'', the images synthesized with the regularization term produce an improved correlation compared to random noise, which, however, is still not sufficient for performing architecture search, as shown in Fig.~\ref{fig:consistency_imagenet} (b). The proposed recursive label calibration and regional update scheme further improve the correlation with $\rho$ increasing from 0.87 to 0.96, in Fig.~\ref{fig:consistency_imagenet} (d), showing that the images synthesized with the proposed method produce a more correlated architecture ranking, which better guarantees the architecture searched on the synthesized dataset will achieve high accuracy on the target dataset.

\vspace{-0.1in}
\subsection{Search Results}
\label{sec:nas_results}
We investigate the effectiveness and generality of our data-free neural architecture search framework by testing it with three prevalent neural architecture search algorithms: DARTS~\cite{liu2018darts}, SPOS~\cite{guo2019single} and ProxylessNAS~\cite{cai2018proxylessnas}.

\begin{table}[t]
\centering
\resizebox{0.75\textwidth}{!}{
\setlength\tabcolsep{2pt}
\begin{tabular}{lcccccc}
\hline
\hline
\noalign{\smallskip}
Methods & Top-1 Err (\%) & Params (M) & \hspace{-1.25em} Data for NAS\\
\noalign{\smallskip}
\hline
\noalign{\smallskip}
NASNet-A$^\dagger$~\cite{zoph2018learning} & $2.65$ & $3.3$ & \hspace{-1.25em} CIFAR-10 \\
BlockQNN~\cite{zhong2018blockqnn} & $3.54$ & $39.8$ & \hspace{-1.25em} CIFAR-10 \\
AmoebaNet-A$^\dagger$~\cite{real2018regularized} & $3.12$ & $3.1$ & \hspace{-1.25em} CIFAR-10 \\
PNAS~\cite{liu2018progressive} & $3.41\pm0.09$ & $3.2$ & \hspace{-1.25em} CIFAR-10 \\
ENAS$^\dagger$ ~\cite{pham2018efficient} & $2.89$ & $4.6$ & \hspace{-1.25em} CIFAR-10 \\
\noalign{\smallskip}
\hline
\noalign{\smallskip}
Random search$^\dagger$~\cite{liu2018darts} & $3.29\pm0.15$ & $3.2$ & \hspace{-1.25em} CIFAR-10\\
DARTS$^\dagger$~\cite{liu2018darts} & $2.76\pm0.09$ & $3.3$ & \hspace{-1.25em} CIFAR-10\\
\noalign{\smallskip}
\hline
\noalign{\smallskip}
\textbf{Data-free DARTS} & $\mathbf{2.68}$ $\pm$ $\mathbf{0.09}$ & $3.3$ & \hspace{-1.25em} \textbf{Synthesized data}\\
\noalign{\smallskip}
\hline
\hline
\noalign{\smallskip}
\end{tabular}}
\caption{Comparison between data-free DARTS and other NAS algorithms on CIFAR-10. $\dagger$ denotes using cutout.}
\label{table:NAS_darts_cifar10}
\end{table}

\begin{table}[t]
\centering
\resizebox{0.75\textwidth}{!}{
\setlength\tabcolsep{2pt}
\centering
\begin{tabular}{lcccccc}
\hline
\hline
\noalign{\smallskip}
Methods & Top-1 Err (\%) & Params (M) & Data for NAS \\
\noalign{\smallskip}
\hline
\noalign{\smallskip}
DARTS~\cite{liu2018darts} & $26.7$ & $4.7$ &  ImageNet\\
\noalign{\smallskip}
\hline
\noalign{\smallskip}
\textbf{Data-free DARTS} & $\mathbf{26.4}$ & $4.7$ & \textbf{Synthesized data}\\
\noalign{\smallskip}
\hline
\hline
\noalign{\smallskip}
\end{tabular}
}
\caption{Generalization of architecture searched with data-free DARTS~\cite{liu2018darts} from CIFAR-10 to ImageNet.}
\vspace{-1em}
\label{table:NAS_darts_ImageNet}
\end{table}

\begin{table}[t]
\centering
\resizebox{0.75\textwidth}{!}{
\setlength\tabcolsep{2pt}
\centering
\begin{tabular}{lcccccc}
\hline
\hline
\noalign{\smallskip}
Methods & \hspace{-1.5em} Top-1 Err (\%) & FLOPs (M) & Data for NAS \\
\noalign{\smallskip}
\hline
\noalign{\smallskip}
all choice \_3 & \hspace{-1.5em} $26.6$ & $324$ & ImageNet \\
all choice \_5 & \hspace{-1.5em} $26.5$ & $321$ & ImageNet \\
all choice \_7 & \hspace{-1.5em} $26.4$ & $327$ & ImageNet \\
all choice \_x & \hspace{-1.5em} $26.5$ & $326$ & ImageNet \\
\noalign{\smallskip}
\hline
\noalign{\smallskip}
Random Select & \hspace{-1.5em} $\sim 26.3$ & $\sim 320$ & ImageNet\\
Random Search & \hspace{-1.5em} $26.2$ & $323$ & ImageNet\\
\noalign{\smallskip}
\hline
\noalign{\smallskip}
SPOS~\cite{guo2019single} & \hspace{-1.5em} $\mathbf{25.7}$ & $319$ & ImageNet\\
\noalign{\smallskip}
\hline
\noalign{\smallskip}
\textbf{Data-free SPOS} & \hspace{-1.5em}  $25.8$ & $316$ & \textbf{Synthesized data}\\ \noalign{\smallskip}
\hline
\hline
\noalign{\smallskip}
\end{tabular}
}
\caption{Comparison between data-free SPOS, original SPOS~\cite{guo2019single} as well as other baseline results. ``All choice'' refers to the baseline algorithm where the same operation is chosen for all layers. ``All Choice \_3, \_5, \_7'' denotes choosing only a ShuffleNet~\cite{zhang2018shufflenet} block with $3\times3$, $5\times5$, or $7\times7$ convolution, respectively. ``Choice \_x'' denotes using the Xception block~\cite{chollet2017xception}.} \label{table:NAS_SPOS}
\end{table}

\begin{table}[t]
\centering
\resizebox{0.75\textwidth}{!}{
\setlength\tabcolsep{2.1pt}
\begin{tabular}{lcccccc}
\hline
\hline
\noalign{\smallskip}
Methods & \hspace{-1.5em} Top-1 Err (\%) & FLOPs(M) & \hspace{-1em} Data for NAS \\
\noalign{\smallskip}
\hline
\noalign{\smallskip}
ProxylessNAS~\cite{cai2018proxylessnas} &  \hspace{-1.5em} $24.4$ & $467$  &  \hspace{-1em} ImageNet\\
\noalign{\smallskip}
\hline
\noalign{\smallskip}
\textbf{Data-free ProxylessNAS} & \hspace{-1.5em} $\mathbf{24.2}$ & $465$ & \hspace{-1em} \textbf{Synthesized data}\\
\noalign{\smallskip}
\hline
\hline
\noalign{\smallskip}
\end{tabular}
}
\caption{Comparison between RL-based ProxylessNAS integrated with our data-free NAS framework and the original ProxylessNAS. We choose the targeting metrics as FLOPs.}
\label{table:NAS_Proxyless}
\end{table}

\setlength{\tabcolsep}{3pt}
\begin{table}[t]
\centering
\resizebox{0.72\textwidth}{!}{
\centering
\begin{tabular}{ccccccccc}
\noalign{\smallskip}
\hline
\hline
\noalign{\smallskip}
Methods & Data Type & Top-1 Err (\%) & FLOPs (G) \\
\noalign{\smallskip}
\hline
\noalign{\smallskip}
ImageNet & Original & $23.4$ & 4.1 \\
\noalign{\smallskip}
\hline
\noalign{\smallskip}
BigGAN~\cite{brock2018large} & GAN synthesized & $37.0$ & $\sim$1.2\\
DI~\cite{yin2020dreaming} & Synthesized & $44.1$ & $\sim$1.2 \\
ADI~\cite{yin2020dreaming} & Synthesized & $39.3$ & $\sim$1.2 \\
\noalign{\smallskip}
\hline
\noalign{\smallskip}
\textbf{Ours} & Synthesized & $\mathbf{36.5}$ & $\mathbf{1.0}$ \\
\noalign{\smallskip}
\hline
\hline
\end{tabular}}
\vspace{0.5em}
\caption{Accuracy comparison for data-free pruning.}
\label{table:pruning}
\end{table}
\setlength{\tabcolsep}{3pt}

\setlength{\tabcolsep}{3pt}
\begin{table}[t]
\centering
\resizebox{0.36\textwidth}{!}{
\centering
\begin{tabular}{ccccccccc}
\noalign{\smallskip}
\hline
\hline
\noalign{\smallskip}
Methods & Top-1 Err (\%) \\
\noalign{\smallskip}
\hline
{Noise} & $86.39$ \\
DeepDream~\cite{mordvintsev2015inceptionism} & $70.02$\\
DAFL~\cite{chen2019data} & $7.78$\\
ADI~\cite{yin2020dreaming} & $6.74$ \\
Ours & $\mathbf{5.97}$\\
\hline
\hline
\end{tabular}}
\vspace{0.5em}
\caption{The top-1 error of data-free knowledge transfer using ResNet-34 as the pre-trained teacher model for synthesize images and train a ResNet-18 student model from scratch for CIFAR-10. }
\label{table:cifar10_transfer}
\end{table}

\setlength{\tabcolsep}{2.5pt}
\begin{table}[t]
\centering
\resizebox{0.75\textwidth}{!}{
\centering
\begin{tabular}{ccccccccc}
\noalign{\smallskip}
\hline
\hline
\noalign{\smallskip}
Methods & Data Type & Data Amount & Epochs & Top-1 Err (\%)\\
\noalign{\smallskip}
\hline
\noalign{\smallskip}
ImageNet & Original & $1.3$M & 250 & $23.4$ \\
\noalign{\smallskip}
\hline
\noalign{\smallskip}
BigGAN~\cite{brock2018large} & GAN synthesized & $215$K & 90 & $36.0$ \\
ADI~\cite{yin2020dreaming} & Synthesized & $140$K & 250 & $26.2$ \\
\noalign{\smallskip}
\hline
\noalign{\smallskip}
\textbf{Ours} & Synthesized & $140$K & 250 & $\mathbf{25.9}$ \\
\textbf{Ours} & Synthesized & $140$K & 500 & $\mathbf{24.8}$ \\
\noalign{\smallskip}
\hline
\hline
\end{tabular}}
\vspace{0.5em}
\caption{Data-free knowledge transfer on ImageNet from pre-trained ResNet-50 to the same network initialized from scratch.}
\label{table:imagenet_transfer}
\end{table}

\begin{figure*}[t]
\begin{minipage}[h]{.62\linewidth}
\includegraphics[width=1.05\linewidth]{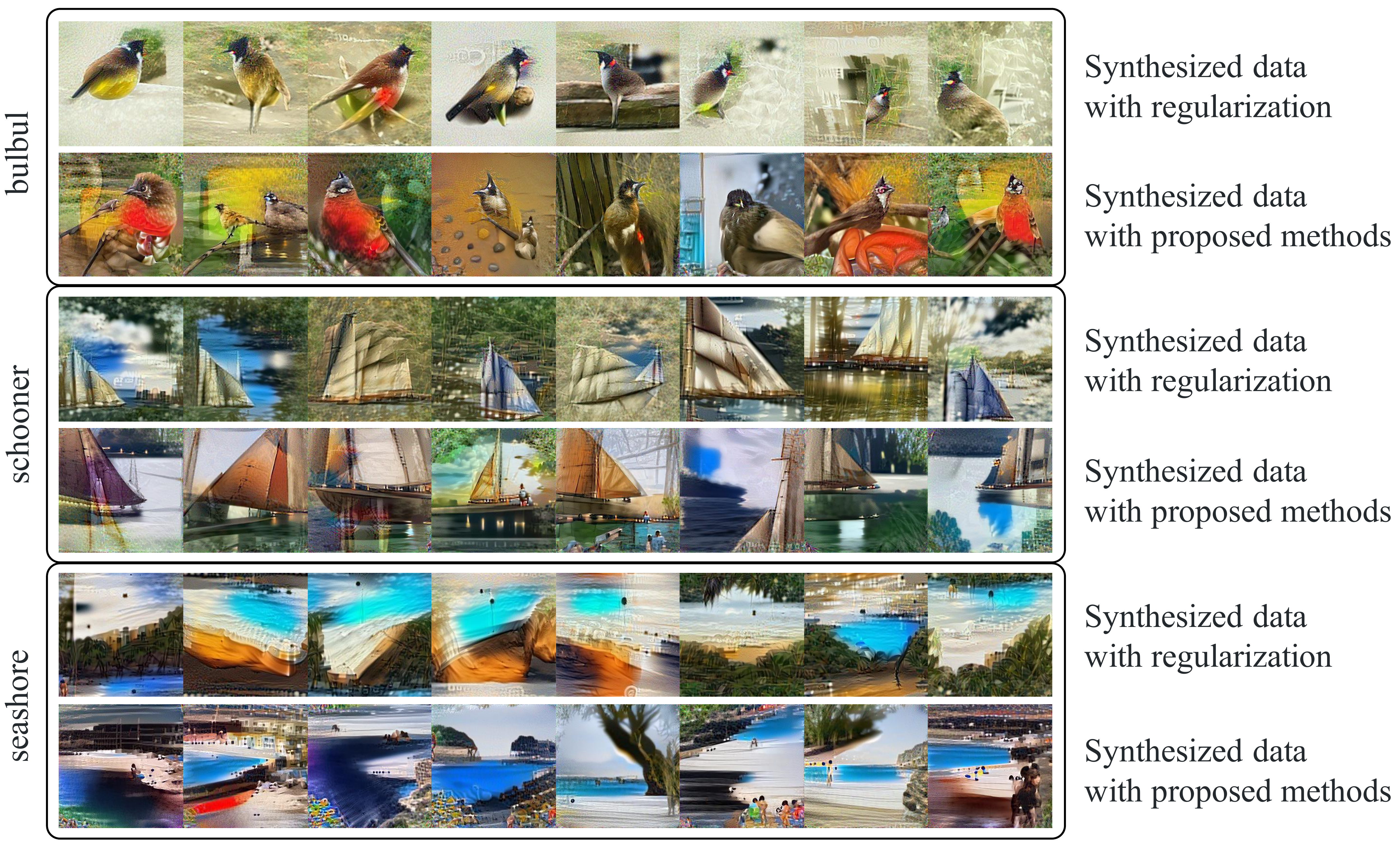}
\caption{Synthesized 256$\times$256 images with ResNet-50 optimized on ImageNet as the pre-trained network. We visualize three classes (\textit{``blublu''},\textit{``schooner''} and \textit{``seashore''}) with different data synthesis methods.}
     \label{fig:synthesized_images}
\end{minipage}
\hfill
\begin{minipage}[h]{.34\linewidth}
\centering
\includegraphics[width=0.64\linewidth]{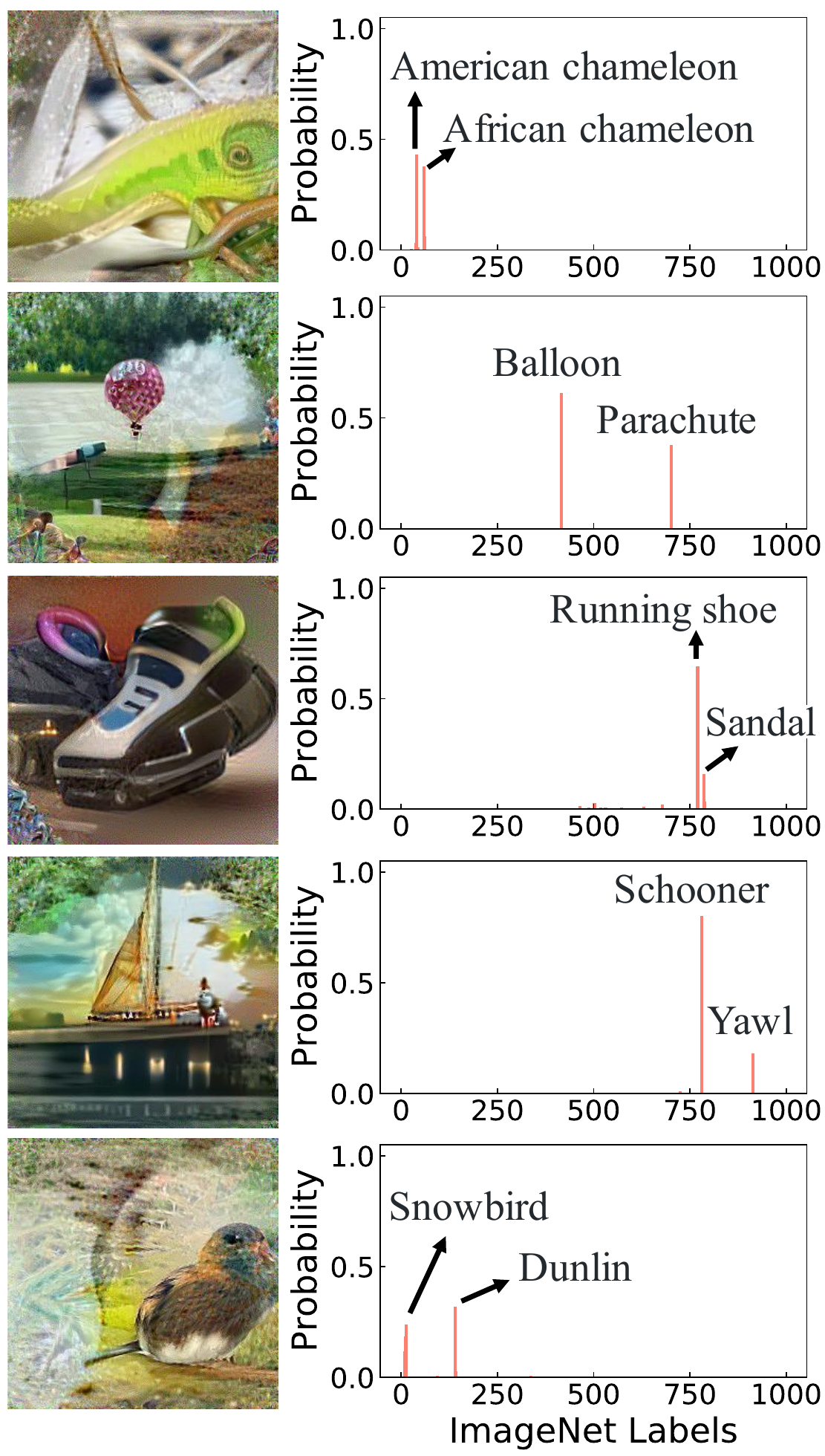}

\caption{Recursive label calibration automatically learns semantically-related labels for the synthetic images.}
\label{fig:visualize_label}
\end{minipage}
\end{figure*}

\vspace{0.6em}
\noindent{\textbf{Implementation Details}}
\vspace{0.2em}

\noindent Our experiments on DARTS are targeted at classification over CIFAR-10~\cite{cifar10} dataset and our experiments on SPOS and ProxylessNAS are targeted at classification over ImageNet~\cite{imagenet} dataset. More experimental details can be referred to our Appendix.

\vspace{0.6em}
\noindent{\textbf{Instantiations}}
\label{sec:search_exp}
\vspace{0.2em}

\noindent{\textbf{Data-free DARTS:}} The architecture found by using DARTS with the data synthesized with the proposed method achieves 2.68\% test error. As shown in Table~\ref{table:NAS_darts_cifar10}, this is comparable to the architecture found when using DARTS with the original CIFAR-10 data. The search result of using data synthesized without label calibration is comparable to random search. This shows that the proposed synthesis method generates data with high correlation to the original CIFAR-10 data, which provides a basis for searching architectures on synthetic data that achieve high performance when evaluated on the original data.
Further, the architecture discovered from the data synthesized with label calibration also generalizes well to the ImageNet dataset and achieves slightly higher accuracy than original DARTS, shown in Table~\ref{table:NAS_darts_ImageNet}.

\noindent\textbf{Data-free Single Path One-Shot:} As shown in Table~\ref{table:NAS_SPOS}, compared to data synthesized without label calibration, the data synthesized with label calibration provides better guidance and results in higher final accuracy. The architecture discovered using synthesized data with label calibration achieves 74.2\% top-1 accuracy. This accuracy is comparable to the results obtained by the SPOS search on the original ImageNet. This result also far exceeds random search as well as the baseline method of choosing the same operation for all layers, demonstrating the feasibility of using synthetic images for data-free NAS on a large-scale dataset.

\noindent\textbf{Data-free ProxylessNAS:} We adopt the RL-based ProxylessNAS targeting at the FLOPs constraint. Table~\ref{table:NAS_Proxyless} shows that ProxylessNAS, when integrated with our data-free framework, also achieves accuracy comparable to ProxylessNAS searching on the original data.

\vspace{-0.1in}
\subsection{Extension Tasks}
\label{sec:extension_exp}
\noindent\textbf{Data-free Pruning:}
\label{sec:metapruning}
We show that the proposed data-free NAS can be applied to pruning tasks via integrating with a search-based pruning method, MetaPruning~\cite{liu2019metapruning}.
We use synthesized ImageNet images to guide MetaPruning for finding the best pruning ratio in each layer and train the searched pruned network from scratch also with the synthesized images. Results in Table~\ref{table:pruning} show that our data-free pruning achieves much higher accuracy than the previous state-of-the-art data-free pruning~\cite{yin2020dreaming}. It further surpasses pruning using GAN-synthesized images. See the appendix for full training details.

\noindent\textbf{Data-free Knowledge Transfer:}
The synthesized images can also serve as a foundation for knowledge transfer from a pre-trained teacher to the student network through training the student network from scratch with knowledge distillation~\cite{hinton2015distilling,shen2020label,muller2019does,shen2019meal} on the synthesized images.
Compared to the previous state-of-the-art data-free learning methods~\cite{yin2020dreaming,chen2019data}, we achieve higher knowledge transfer accuracy when the target dataset is CIFAR-10 or ImageNet, as shown in Table~\ref{table:cifar10_transfer} and \ref{table:imagenet_transfer}. Full training details are included in the appendix.

\subsection{Visualization}
\label{sec:visualization}
In Fig.~\ref{fig:synthesized_images}, we can observe that images synthesized with regularization terms only have homogeneous color palettes and similar backgrounds, but when using the proposed label calibration and regional update scheme for synthesis, the images exhibit more diverse feature and look more realistic. Fig.~\ref{fig:visualize_label} further shows a promising finding that the images synthesized with recursive label calibration do capture the underlining class relationships in a similar manner to the natural images. These attributes produce a good accuracy correlation between architectures trained on synthesized images and natural images which in turn contributes to the success of data-free NAS.

\section{Conclusion}

We have presented a novel framework that can conduct neural architecture search (NAS) without access to the original data. In order to perform data-free NAS, we developed a new recursive label calibration method and regional update strategy which automatically discover and encode the relationships between classes in soft labels and generate images \textit{w.r.t.} these soft labels, while preserving the diversity of generated images. We demonstrate the effectiveness of our proposed method with three typical NAS algorithms. Our experimental results show that using the recovered data from a pre-trained model, NAS algorithms using our data synthesis method can obtain performance comparable to NAS algorithms using the original training data. This verifies that it is feasible to conduct NAS without original data if the synthesis method is well designed. In addition, we demonstrate that our synthetic data can produce state-of-the-art results for data-free pruning as well as knowledge transfer.

\clearpage

\bibliographystyle{splncs04}
\bibliography{main}

\appendix
\section*{Appendix}
\section{Details of Consistency Exploration}

For DARTS search space, we randomly sample 100 architectures, and train each architecture from scratch using original CIFAR-10 data, synthetic data, or random noise separately and report the architecture's accuracy when evaluated on original CIFAR-10 validation dataset.
When training on original CIFAR-10, we use cross-entropy loss. For the synthesized data and random noise, we use the KL-divergence loss between the network's output and the soft labels.

We use the SPOS~\cite{guo2019single} search space to explore the more challenging task of architecture search on the large-scale ImageNet dataset.
We randomly sample 1000 architectures from the SPOS search space, then evaluate their accuracy. Since training thousands of architectures from scratch on the ImageNet dataset is computationally prohibitive, we train a SuperNet, which is originally proposed in SPOS, on different data sources including random noise, synthesized data, and ImageNet. We then evaluate the accuracy of 1000 randomly-sampled architectures on each data source. Since the SuperNet contains all the sub-networks and SPOS search algorithm itself uses the accuracy inferred from the SuperNet to rank different architectures in the search stage, a good correlation between the accuracy inferred from the SuperNet trained on synthetic data and the SuperNet trained on ImageNet provides a solid foundation for whether SPOS can discover high-quality architectures using synthetic data.

\section{Training Details of Data-free NAS}

Our experiments on DARTS are targeted at classification over CIFAR-10~\cite{cifar10} dataset and our experiments on SPOS and ProxylessNAS are targeted at classification over ImageNet~\cite{imagenet} dataset.

\subsection{Data synthesis}
For synthesizing data targeting the CIFAR-10 dataset, we use ResNet-34 trained on the CIFAR-10 data as the pre-trained model. We generate 50k 40$\times$40 images, with a batch-size of 250 images per generation. We generate the images for each batch by running our optimization procedure for 2000 iterations in inner loop and 10 iterations in outer loop. We use Adam optimizer with a learning rate of 0.1.
When targeting the ImageNet dataset, we use ResNet-50 trained on ImageNet as the pre-trained model. Then we generate 140k 256$\times$256 images, with a batch-size of 50 images per generation, we use Adam optimizer with learning rate 0.25 to optimize each batch for 5000 iteration in inner loop and 10 iterations in outer loop.
As mentioned in~\cite{liu2020labels}, the primary goal of NAS is to identify a high-quality network architecture. Training the weights of that architecture is not necessarily the objective of a NAS algorithm, which instead, is done during a separate evaluation phase. Thus, to testify that the data-free NAS can discover an architecture that performs well when trained and tested on the target data, and for a fair comparison with other state-of-the-art NAS algorithms, we use the original data for evaluating the searched architectures. We train the searched architecture from scratch with the same data and training schemes as the instantiation NAS methods originally used.

\subsection{Data-free DARTS}
\label{sec:darts_detail}

\noindent\textbf{Search Space}
The DARTS search space is defined as a cell-based search space, where the search process finds cells that can be applied to form a network of $L$ cells. Each cell is a directed acyclic graph (DAG) of $N$ nodes, where each node is a latent representation. The search algorithm searches for the operation of type $o$ between two nodes from the operation space $\mathcal{O}$. The operation space $\mathcal{O}$ consists of: 3$\times$3 and 5$\times$5 separable convolutions, 3$\times$3 and 5$\times$5 dilated separable convolutions, 3$\times$3 max pooling, 3$\times$3 average pooling, identity, and \textit{zero}. More details are described in the original DARTS paper~\cite{liu2018darts}.

\noindent\textbf{Searching Phase}
We use synthetic CIFAR-10 data for data-free DARTS search. It contains 50k 40$\times$40 images, \textit{i.e.}, 5k images in each class. We use random crop augmentation. In each iteration, a 32$\times$32 region is selected to be the input to the DARTS algorithm. The KL-divergence loss is imposed between the output logits of DARTS and the soft-label of the synthetic images for training data-free DARTS. A DARTS convolution cell consists of 7 nodes and a network of 8 cells. The initial number of channels is set to 16. DARTS alternatively updates the architecture parameters and weight parameters. In training data-free DARTS, we follow the training hyper-parameters in original DARTS paper~\cite{liu2018darts} exactly. We hold out half of the synthetic training data as the validation set for search. For the weight parameters we use SGD with momentum 0.9, batch size 64, initial learning rate 0.025, and weight decay 3$\times$10$^{-4}$. The weights are trained for 50 epochs. For architecture variables we use zero initialization and Adam with initial learning rate 3$\times$10$^{-4}$ and weight decay 10$^{-3}$.

\noindent\textbf{Evaluation Phase}
In the evaluation phase, following~\cite{liu2018darts}, the network consists of 20 searched cells and 36 initial channels. It is trained for 600 epochs with batch size 96. Other hyper-parameters are the same as the hyper-parameters used in the search phase.

\subsection{Data-free SPOS}
\label{sec:spos_detail}

\noindent\textbf{Search Space}
The Single Path One-Shot (SPOS)~\cite{guo2019single} search space is a block-wise search space where each block can choose a different block $o$ from the operation space $\mathcal{O}$. The operation space $\mathcal{O}$ contains 4 candidates: the ShuffleNet v2~\cite{zhang2018shufflenet} block with $3\times3$, $5\times5$, or $7\times7$ convolution, and the Xception block~\cite{chollet2017xception}. For more technical details, please refer to the original SPOS paper~\cite{guo2019single}

\noindent\textbf{SuperNet Training Phase}
We use our synthetic ImageNet data for data-free SPOS. We randomly sample 32 synthetic images from each of the 1000 classes to form the validation dataset, which is used for validation during the evolutionary search. The SuperNet is trained using the remaining 108k synthetic images. We use the KL-divergence loss between SuperNet output logits and the soft-labels to train the SuperNet. The training scheme follows SPOS~\cite{guo2019single}. We train the SuperNet for 120 epochs using SGD with momentum 0.9, weight decay 4$\times$10$^{-5}$, batch size 1024, and initial learning rate 0.5.

\noindent\textbf{Searching and Evaluation Phase}
We conduct the evolutionary search with the accuracy of candidate architectures inferred on the synthetic validation set using the weights of the trained SuperNet. Then in evaluation, we train the best architecture from the search phase from scratch. We train for 240 epochs and all other training hyper-parameters are the same as used in training the SuperNet.

\subsection{Data-free ProxylessNAS}
\label{sec:proxylessnas_detail}

\noindent\textbf{Search Space}
The ProxylessNAS search space is based on a MobileNet V2~\cite{sandler2018mobilenetv2} backbone. It is a block-wise search space, searching for a mobile inverted bottleneck convolution (MBConv) in each block of the backbone. The candidate operations are chosen among MBConvs with various kernel sizes \{3, 5, 7\} and expansion ratios \{3, 6\}, as detailed in the original ProxylessNAS paper~\cite{cai2018proxylessnas}

\noindent\textbf{Searching and Evaluation Phase}
ProxylessNAS~\cite{cai2018proxylessnas} trains an over-parameterized network that contains all candidate paths and uses architecture parameters to learn to select among the paths. When training the over-parameterized network, we form a validation set by randomly holding out $\sim$32k images from the synthetic training dataset. Instead of using the original loss, we impose the KL-divergence loss between the output logits of the over-parameterized network and the soft-labels of the synthetic images.
We use the same hyper-parameter settings as the RL-based ProxylessNAS, which uses Adam optimizer with a 0.001 learning rate for training the architecture parameters. Then in evaluation, the searched network is trained from scratch with Adam optimizer for 300 epochs.

\section{Training Details of Extension Tasks}

\subsection{Data-free Pruning}
\label{sec:pruning_detail}
Network channel pruning has been recognized as an effective network compression technique~\cite{he2017channelpruning,liu2017learning,rethinkpruning}. Traditional pruning requires original training data to perform pruning. In this study, we conduct pruning without the original training data by integrating our data-free NAS with a search-based pruning method named MetaPruning~\cite{liu2019metapruning} and show the generalization ability of the proposed data-free NAS framework to the pruning tasks.

\noindent\textbf{Data Preparation:} We use synthesize 140k images generated from a ResNet-50 network pre-trained on the ImageNet dataset. We randomly split this synthetic data into two parts: 32k images for validation during search (\textit{i.e.}, 32 images for each class) and the remaining 108k images are used as the training set.

\noindent\textbf{Pruning Framework:} The optimization framework for training PruningNet, evolutionary search, and evaluating the searched network follows MetaPruning. We first train a ResNet-50-based PruningNet from scratch on the synthetic training set. PruningNet is proposed in MetaPruning to generate weights for the pruned structures. We use the KL-divergence loss between the PruningNet output logits and the soft labels. Then we infer the accuracy of the candidate pruned networks from the synthetic validation set with the weights generated by PruningNet and use evolutionary search to find the best pruned network. After that, we train the best pruned network from scratch using the 140k synthetic images and the corresponding soft labels.

\noindent\textbf{Hyper-parameters:} We adopt the same training hyper-parameters as~\cite{liu2019metapruning}. We randomly crop a $224\times224$ region from the $256\times256$ images and train the PruningNet for 32 epochs with an initial learning rate of 0.1 and batch-size of 256. We use SGD with a momentum of 0.9 and weight decay of $1\times10^{-4}$. In training the pruned network from scratch, we again use the synthetic images. Hyper-parameter settings remain the same as those used for training the PruningNet, except the number of training epochs is set to 128.

\subsection{Data-free Knowledge Transfer}
\label{sec:knowledge_transfer_detail}

Further, we study knowledge transfer from teacher network to student network without using any original data. For a fair comparison with Dreaming to Distill~\cite{yin2020dreaming}, we use a pre-trained ResNet-50 network as the teacher network for image synthesis and train another randomly initialized ResNet-50 from scratch using 140k images synthesized with our proposed methods. We use the synthesized images and the teacher network's predictions on the images as soft-labels to train the student network and train using SGD with momentum 0.875, weight decay 3$\times$10$^{-5}$, batch-size 1024 and initial learning rate 1.024.

\end{document}